\pdfoutput=1
\documentclass[11pt]{article}

\usepackage[]{ACL2023}

\usepackage{times}
\usepackage{latexsym}

\usepackage[T1]{fontenc}
\usepackage[utf8]{inputenc}

\usepackage{microtype}

\usepackage{amsmath}
\usepackage{hyperref}

\usepackage{inconsolata}
\usepackage{graphicx}

\usepackage{placeins}

\usepackage[colorinlistoftodos]{todonotes}

\usepackage{enumitem,kantlipsum}

\title{Evaluating the Quality of Benchmark Datasets for Low-Resource Languages: A Case Study on Turkish}

\author{
   Ayşe Aysu Cengiz$^{1*}$ \,\,
   Ahmet Kaan Sever$^{2*}$   \,\,
   Elif Ecem Ümütlü$^{1*}$ \,\,
   \textbf{Naime Şeyma Erdem}$^3$   \\
   \textbf{Burak Aytan}$^{3}$ \,\,
   \textbf{Büşra Tufan}$^{4}$ \,\,
   \textbf{Abdullah Topraksoy}$^{5}$ \,\,
   \textbf{Esra Darıcı}$^{6}$ \,\,
   \textbf{Cagri Toraman}$^{1}$ \\
   $^1$Middle East Technical University, Computer Engineering Department\\
   $^2$Bilkent University, Computer Engineering Department\,\,
   $^3$Turkcell AI \\
   $^4$Hacettepe University, Sociology Department\,\,
   $^5$Istanbul University, Linguistics Department \\
   $^6$Middle East Technical University, Turkish Language Department \,\, *Equal contribution \\
   \small $^1$\texttt{\{e2580371, e2448991, ctoraman\}@ceng.metu.edu.tr} \,\,
   \small$^2$\texttt{kaan.sever@ug.bilkent.edu.tr} \\ \small$^3$\texttt{\{burak.aytan,seyma.erdem\}@turkcell.com.tr} \,\,
   \small$^4$\texttt{busratufan@hacettepe.edu.tr} \\ 
   \small$^5$\texttt{abdullah.topraksoy@istanbul.edu.tr} \,\, 
   \small$^6$\texttt{darici@metu.edu.tr} 
}

\begin{document}
\maketitle
\begin{abstract}

\end{abstract}
The reliance on translated or adapted datasets from English or multilingual resources introduces challenges regarding linguistic and cultural suitability. This study addresses the need for robust and culturally appropriate benchmarks by evaluating the quality of 17 commonly used Turkish benchmark datasets. Using a comprehensive framework that assesses six criteria, both human and LLM-judge annotators provide detailed evaluations to identify dataset strengths and shortcomings.

Our results reveal that 70\% of the benchmark datasets fail to meet our heuristic quality standards. The correctness of the usage of technical terms is the strongest criterion, but 85\% of the criteria are not satisfied in the examined datasets. Although LLM judges demonstrate potential, they are less effective than human annotators, particularly in understanding cultural common sense knowledge and interpreting fluent, unambiguous text. GPT-4o has stronger labeling capabilities for grammatical and technical tasks, while Llama3.3-70B excels at correctness and cultural knowledge evaluation. Our findings emphasize the urgent need for more rigorous quality control in creating and adapting datasets for low-resource languages.

\section{Introduction}
Natural language processing has made significant advances in recent years, with large language models achieving impressive results in various tasks \cite{srivastava2022beyond, bubeck2023sparks}. However, the quality and reliability of these models mostly depend on the datasets used for training and evaluation \cite{tedeschi2023s}. 

For languages with relatively low resources, such as Turkish, the availability of high-quality datasets is crucial for developing robust and accurate systems. Turkish natural language processing resources are significantly based on datasets translated from English or adapted from multilingual resources \cite{hu2020xtreme,liang-etal-2020-xglue,kuisai2024cetvel}. Although these datasets enable progress in low-resource natural language processing, their quality and suitability for specific tasks are not thoroughly examined. The use of translated or adapted datasets raises concerns about their adherence to grammar, cultural nuances, and overall coherence, potentially leading to biased or inaccurate model performance. %This is mostly due to the unique linguistic features of the Turkish language (e.g., agglutinative morphology) that cause translations to degrade in quality or reflect biases.

Motivated by the need for reliable and culturally appropriate benchmarks in natural language processing, this study aims to evaluate the quality of commonly used data resources in Turkish as a case study. This evaluation is crucial to advance the field of low-resource natural language processing by identifying potential shortcomings.

To address this gap, we present a comprehensive analysis of 17 widely used Turkish datasets. Our rationale for selecting these datasets is that they are widely used datasets in the literature or published within popular benchmarks (see Table \ref{tab:benchmark-data-size}). Our evaluation framework focuses on six key aspects, including answer correctness, grammatical correctness, cohesion and coherence, comprehensibility and fluency, technical term usage, and alignment with cultural common sense. These aspects reflect quality by correctness, grammar capability, and cultural sensitivity. They are designed by domain experts who are also co-authors of this study. A wide range of human annotators manually label samples from each dataset according to these criteria to provide a detailed assessment of their quality and suitability for our target language. We also examine the LLM-as-a-Judge approach \cite{zheng2023judging} to compare its labeling performance with human annotations.

Our findings reveal that 70\% of the benchmark datasets fail to meet our criteria, and 85\% of the criteria are not satisfied by these datasets. LLM judges are not as effective as human annotators, particularly in understanding cultural common sense knowledge, and interpreting fluent and unambiguous text. Our results emphasize the importance of developing high-quality and novel benchmark datasets for more accurate and culturally sensitive settings. The observations are valuable not only for the Turkish language but also for all languages that need high-quality data resources in terms of correctness, grammar, and cultural sensitivity\footnote{We publish all related material including data, annotation details, scripts, and prompts online at https://github.com/metunlp/llmevaluation}.

\section{Related Work}
\paragraph{Dataset Quality}
Dataset quality is assessed by different methods in the literature. \citet{kreutzer2022quality} sampled 100 instances from each dataset, as in our study, to identify the data quality of multilingual web-crawled datasets. Their findings reveal that many datasets suffered from quality issues, primarily due to the nature of web crawling. 

The GSM1k dataset \cite{zhang2024careful} evaluates the performance of language models on reasoning tasks. The dataset is kept private to prevent contamination. They conducted a three-stage annotation process that includes an initial review by experienced annotators, a secondary validation by independent annotators, and a final audit by a dedicated quality assurance team. 

\paragraph{Contamination}
Data contamination in large language models has become an increasing concern. As models are trained on large-scale datasets scraped from the Internet, the integrity of benchmark datasets is challenging to maintain. \citet{sainz2023did} emphasize the critical need to assess whether a model's performance is due to its genuine reasoning capabilities or mere memorization.

Contamination is detected by matching test splits with training data. \citet{dodge2021documentinglargewebtextcorpora} employ exact match detection methods, normalizing text for capitalization and punctuation to identify instances of overlap. \citet{brown2020languagemodelsfewshotlearners}, on the other hand, use n-gram overlap to measure contamination.

The MEGA benchmark \cite{ahuja2023megamultilingualevaluationgenerative} has a comprehensive case study on contamination by detecting potential training data leakage. They show that some of the benchmark datasets, which were translated into Turkish and analyzed in this study, exhibit data contamination.

\paragraph{Annotation Guideline}
Several studies have established guidelines for human evaluation to ensure consistency and reliability. \citet{liang2023holisticevaluationlanguagemodels} emphasized the importance of structured annotation guidelines to provide a clear and replicable evaluation criteria.

\citet{liang2023holisticevaluationlanguagemodels} designed annotation guidelines to assess disinformation scenarios. To maintain annotation reliability, they implemented quality control measures including hidden “secret words” in instructions to verify comprehension and attention checks to detect careless responses. %They also ensured replicability in their human evaluations by structuring their guidelines with clear criteria, explicit rating scales, and quality control mechanisms.

\paragraph{LLM-as-a-Judge}
\citet{zheng2023judging} propose the method of using powerful LLMs to label and score from a group of candidates. \citet{bavaresco2024llms} introduced the Judge-Bench, a benchmark that evaluates LLM's abilities to replicate human judgments. This benchmark incorporates 20 diverse datasets, each focusing on different tasks and annotation methods. Their findings reveal that while LLMs can effectively align with human judgments in specific tasks, their performance varies significantly across different tasks.

\citet{verga2024replacing} proposed that using a smaller group of LLMs together, called LLM Jury, instead of relying on a single model would yield a higher correlation with human evaluation. This alternative approach reduces costs while improving reproducibility and applicability. %Their findings indicate that combining evaluations from a diverse set of models helps minimize intra-model scoring bias.

\citet{srivastava2022beyond} introduced BIG-bench, a collaborative benchmark comprised of 204 diverse tasks designed to evaluate the capabilities of LLMs. Their findings show that large models struggle with tasks that require complex reasoning and understanding, and LLM performance is relatively worse than human annotators.

\paragraph{Our Differences} 

This study evaluates the quality of LLM benchmarks in a comprehensive framework that includes multiple criteria. We conduct a use case study on popular Turkish datasets for this framework. The approach described in this study can be generalized to other languages that suffer from having low resources and cultural sensitivity.

\section{Datasets}
This study examines 17 datasets, listed in Table \ref{tab:benchmark-data-size}. We provide the details of each dataset, as follows.

\paragraph{XQuAD}
XQuAD \cite{artetxe2019cross} is a multilingual open-ended reading comprehension benchmark. The dataset includes 1.190 question-answer pairs from the SQuAD v1.1 benchmark \cite{rajpurkar2016squad} and human translations from the English text to 10 different languages including Turkish. The model is evaluated based on its capability to extract correct answers from a given passage.

\paragraph{XCOPA}
XCOPA \cite{ponti2020xcopa} is a multilingual dataset of common sense causal reasoning. XCOPA is the human translation of the verification and test sets of COPA \cite{roemmele2011choice} to 11 languages including Turkish. This dataset evaluates the model based on its understanding of causal relations and inferential capability.

\paragraph{Belebele}
Belebele \cite{bandarkar2023belebele} is a multiple-choice and multilingual reading comprehension benchmark. The multilingual passages are obtained from Flores-200 \cite{nllb2022}, and the questions were written by humans. The benchmark was translated from English into other languages including Turkish, resulting in a 122 language multilingual dataset. Belebele evaluates the LLM model's understanding of the information given in the text.

\paragraph{XL-Sum}
XL-Sum \cite{hasan2021xl} is a multilingual summarization benchmark. The dataset spanning 44 languages was created with a similar process as XSUM \cite{narayan2018don}. In addition, the quality of the summaries in 10 languages were evaluated by human annotators. This benchmark aims at abstractive summarization in which the summary can have new phrases that are not present within the original text. 

\begin{table}[t!]
\small
\begin{tabular}{|l|l|l|l|l|}
\hline
\textbf{Dataset} &
  \textbf{Size} &
  \textbf{Cite} &
  \begin{tabular}[c]{@{}l@{}}\textbf{Dload}\end{tabular} &
  \textbf{Bench.}\\ \hline
XQuAD                                                                 & 1.190   & 791 & 5k  & \begin{tabular}[c]{@{}l@{}}XTREME\\ MEGA\\ Cetvel\end{tabular} \\ \hline
XCOPA                                                                 & 600    & 250 & 6k   & \begin{tabular}[c]{@{}l@{}}MEGA\\ Cetvel\end{tabular}          \\ \hline
Belebele                                                              & 900    & 79  & 14k  & Cetvel                                                         \\ \hline
XL-Sum                                                                & 34k    & 365 & 114k & \begin{tabular}[c]{@{}l@{}}
    MEGA \\
    Cetvel
\end{tabular}                                                      \\ \hline
XNLI &
  400.2k &
  1.4k &
  14k &
  \begin{tabular}[c]{@{}l@{}}XTREME\\ MEGA\\ XGLUE\\ Cetvel\end{tabular} \\ \hline
\begin{tabular}[c]{@{}l@{}}Turkish PLU\\ Linking\end{tabular} & 1.759   & 4   & 48      & Cetvel                                                         \\ \hline
\begin{tabular}[c]{@{}l@{}}Turkish PLU\\ Goal Infer\end{tabular}  & 260.8k & 4   & 213     & Cetvel                                                         \\ \hline
\begin{tabular}[c]{@{}l@{}}Turkish PLU\\ Step Infer\end{tabular}  & 129.6k & 4   & 190     & Cetvel                                                         \\ \hline
\begin{tabular}[c]{@{}l@{}}Turkish PLU\\ Step Ordering\end{tabular}   & 550k   & 4   & 128     & Cetvel                                                         \\ \hline
\begin{tabular}[c]{@{}l@{}}Turkish PLU\\ Next Event \\ Prediction\end{tabular} &
  93k &
  4 &
  130 &
  Cetvel \\ \hline
\begin{tabular}[c]{@{}l@{}}Turkish PLU\\ Summarization\end{tabular}   & 125k   & 4   & -       & Cetvel                                                         \\ \hline
WikiANN                                                               & 40k    & 511 & 63k   & \begin{tabular}[c]{@{}l@{}}XTREME\\ MEGA\end{tabular}          \\ \hline
UDPOS v2.5                                                            & 9.4k   & 142 & -       & \begin{tabular}[c]{@{}l@{}}XTREME\\ MEGA\end{tabular}          \\ \hline
MKQA                                                                  & 10k    & 148 & 284     & Cetvel                                                         \\ \hline
\begin{tabular}[c]{@{}l@{}}OffensEval \\ TR-2020\end{tabular}         & 35.2k  & 177 & 391     & Cetvel                                                         \\ \hline
STS-B-TR                                                         & 8.6k   & -   & 397     & Cetvel                                                         \\ \hline
MMLU-Pro-TR                                                           & 11.9k  & -   & 180     & -                                                              \\ \hline
\end{tabular}
\caption{The details of 17 datasets examined in this study. \emph{Size} refers to the number of total instances, \emph{Cite} refers to the number of citations when this study is published, \emph{Dload} refers to the approximate number of downloads from Huggingface when this study is published, and \emph{Bench} refers to the benchmarks that involve a corresponding dataset (XTREME \cite{hu2020xtreme}, MEGA \cite{ahuja2023megamultilingualevaluationgenerative}, XGLUE \cite{liang-etal-2020-xglue}, and Cetvel \cite{kuisai2024cetvel}). Empty cells mean that dataset does not have a publication, or is not published at Huggingface or in a benchmark.}
\label{tab:benchmark-data-size}
\end{table}

\paragraph{XNLI}
XNLI \cite{conneau2018xnli} is a multilingual natural language inference benchmark. This dataset is obtained from the human translations of MultiNLI \cite{williams2017broad} into 15 languages. Model is evaluated on the basis of their ability to recognize textual entailment. %Given two sentences, a model must determine whether their relationship is one of "entailment", "contradiction", or "neutral".

\paragraph{Turkish PLU}

Turkish PLU \cite{uzunoglu2023benchmarking} is a language understanding benchmark based on Turkish WikiHow, having six subsets as follows. \textbf{Goal Inference} evaluates the model's ability to identify the overarching goal based on a given step. In \textbf{Step Inference}, the model is expected to find the step that needs to be taken to reach a goal. \textbf{Step Ordering}, given a goal and two steps, expects the model to find the preceding step out of the two. In \textbf{Next Event Prediction}, a goal and a step are given, and the model should determine which of the four candidate steps follows the given step. \textbf{Summarization} is an abstractive summarization task. \textbf{Linking Actions} contains WikiHow dump, goal-step matches as the ground-truth, and the dumped steps from WikiHow matched with the goal.

\paragraph{WikiANN}
WikiANN \cite{pan-etal-2017-cross} is a multilingual Named Entity Recognition (NER) dataset that spans more than 282 languages including Turkish. The tagged sentences are directly from Turkish Wikipedia. The benchmark utilizes Wikipedia markups to label PER (person), LOC (location), and ORG (organization) in IOB2 format.

\paragraph{Universal Dependencies v2.5} 
This is a Part of Speech (POS) data from the XTREME benchmark, based on the Universal Dependencies v2.5 tree banks \cite{11234/1-3105} that comprises of multi-lingual POS tagged sentences. 
 
\paragraph{MKQA}
MKQA \cite{longpre-etal-2021-mkqa} is a multilingual question answering benchmark that includes human translates from the English Natural Questions (NQ) \cite{kwiatkowski-etal-2019-natural}, where the questions are obtained from Google queries. The model is evaluated on the basis of their ability to respond correctly to knowledge-based questions. 

\paragraph{OffensEval-TR 2020}
\citet{ccoltekin2020corpus} have sentences extracted from Turkish tweets that are labeled as offensive or non-offensive. The dataset also breaks down the offensive label into two as targeted and not-targeted. Targeted label is further split into group, individual, and other. 

\paragraph{STSb-TR}
STSb-TR \cite{beken-fikri-etal-2021-semantic} is a semantic textual similarity benchmark in Turkish, which is machine-translated from STSb English dataset \cite{cer2017semeval}. Two sentences are given and a decimal score between 0.0 and 5.0 is the target prediction, where a score closer to 5.0 means that the sentences portray more similar meaning.

\paragraph{MMLU-Pro-TR}
MMLU-Pro-TR \cite{MMLU-pro-TR} is the machine translated version of MMLU-Pro \cite{wang2024mmluprorobustchallengingmultitask}, which is the updated version of MMLU \cite{hendrycks2021measuringmassivemultitasklanguage}. The translation is provided by Gemini 1.5 Pro with human oversight. MMLU-Pro-TR also includes hand-picked STEM problems, TheoremQA, and SciBench in addition to MMLU-Pro.

\section{Methods}
In this section, we present our criteria for assessing the quality of datasets. We then explain two types of evaluation; human annotations and LLM-Judge. 

\subsection{Criteria}
\label{sec:criteria}

In order to systematically assess the overall quality and reflectivity of Turkish understanding in all datasets, we establish six distinct criteria. These criteria are designed to ensure a comprehensive evaluation, covering both linguistic precision and cultural understanding.

\paragraph{Answer Correctness}
This criterion assesses whether the dataset’s provided “gold” answer is factually or logically correct for the given prompt or question. An answer is considered correct if it aligns with verified knowledge, is relevant to the question or task, and does not contain incorrectness or information loss due to translation errors or data processing. 

\paragraph{Grammatical Correctness}
This criterion evaluates whether sentences comply with Turkish morphological, orthographic, and syntactic rules. The evaluation is supported by the grammatical rules documented by the linguistic experts, given in Appendix \ref{appendix:annotation_guidelines}. 

\paragraph{Cohesion and Coherence}
This criterion measures both the logical and linguistic completeness of the text. Cohesion is a grammatical, lexical, and semantic issue, based on the fact that linguistic elements do not contradict each other and form a linguistic and semantic integrity. %Cohesion refers to the connection of various parts of a text in a way that ensures linguistic unity \cite{onursal-2004}. It is a concept that enables a writing to be recognized as a text, establishing intra-textual relationships and encompassing all linguistic features \cite{gunay-2007}.

Coherence refers to the logical connection within a text. Consistency emerges by questioning the content expressed in language and its semantic and logical relationship with both the text itself and the realities in the outside world. An entry is considered coherent if the logical relationship between words, sentences, and ideas is clear and well-structured, ensuring that the text has a consistent meaning in its entirety.

\paragraph{Comprehensibility, Fluency, and Ambiguity}
This criterion aims to capture the naturalness of the text, i.e. whether a native speaker would find the sentence clear, smooth, and idiomatic. Ambiguity examines whether the text is ambiguous or vague in a way that prevents a consistent interpretation. Ambiguity evaluation is supported by the ambiguity guidelines documented by the linguistic experts, given in the Appendix. 

\paragraph{Technical and Special Term Usage}
This criterion examines whether domain-specific or technical terms (e.g., legal, medical, or academic) are used or translated accurately. 

\paragraph{Compliance with Cultural Common Sense Knowledge}
Although each individual has common sense knowledge \cite{anacleto2006can}, this knowledge varies from culture to culture and region to region. The model should consider the behaviors and characteristics of specific sociocultural groups \cite{nguyen2023extracting}. This criterion evaluates whether the dataset is in line with the social, economic, cultural, and geographical norms of the language. 

Within the scope of this study, to evaluate the datasets' suitability to Turkish cultural common sense knowledge and ensure that it is comprehensive, the cultural common sense knowledge criteria of different studies are used together \cite{anacleto2006can, shwartz2022good, deshpande2022stereokg, yin2022geomlama}. The following components (food and meal times, drinks, clothing, rituals and traditions, behaviors, social norms, and sports) are dynamics that express common culture, and these dynamics are also determinants of common sense. These judgments vary according to classes, status, beliefs, education levels, gender, race, and ethnicity. Our aim is therefore not to present a definitive scientific survey but to reach reasonable assumptions. In this context, the aim is to bring cultural differences into machine-readable form.

This evaluation is designed by sociologists who are experts in cultural common sense, and based on two main components (details are given in Appendix \ref{appendix:annotation_guidelines}):

\begin{enumerate}[wide, labelwidth=!, labelindent=0pt, label=\roman*.]
    \item \textbf{Contextual Relevance}: The information should accurately reflect Turkey's rules, laws, political structure, and social customs. Data containing foreign legal systems, measurement units, or culturally irrelevant concepts (e.g., feet, inches, gallons) are considered non-compliant.
    
    \item \textbf{Cultural Appropriateness}: This component examines common practices and traditions in Turkey. We adapt different approaches to cover cultural practices and traditions \cite{Nguyen-2023,anacleto2006can,acharya-2020, shwartz2022good, yin2022geomlama}. We particularly examine cultural appropriateness in terms of food and meal, drinks, clothing, rituals and traditions, sports, and social norms.

        %\noindent \emph{Food and Meal Times}: Typical Turkish breakfast, lunch, and dinner items should be accurately represented. Non-Turkish meal habits (e.g., bacon for breakfast, club sandwiches for dinner) indicate non-compliance.
        
        %\noindent \emph{Drinks}: Beverages like Turkish coffee, rakı, ayran, and şalgam are culturally appropriate, while drinks associated with other cultures (e.g., Uzo, Christmas beverages) are not.
        
        %\noindent \emph{Clothing}: Traditional and commonly worn Turkish attire (e.g., şalvar, kaftan, başörtüsü) is considered appropriate, while foreign traditional clothing (e.g., Scottish kilt) is not.
        
        %\noindent \emph{Rituals and Traditions}: Events such as weddings, circumcision ceremonies, and religious holidays should align with Turkish customs. Practices like hand-kissing during holidays or large wedding gatherings are considered culturally appropriate, whereas Western-style wedding receptions or champagne popping are not.
        
        %\noindent \emph{Behaviors and Social Norms}: Politeness towards elders and common social etiquette are expected, while behaviors like public spitting or violating traffic rules are considered non-compliant.
        
        %\noindent \emph{Sports}: Popular sports in Turkey, such as football, wrestling, and swimming, are acceptable, while sports uncommon in Turkish culture, like American football, are not.

\end{enumerate}

\subsection{Evaluation Methods}
\subsubsection{Human Evaluation}
%Human annotation is essential for our work to reliably measure the reflectivity of Turkish understanding in the used datasets. 
Human evaluation is superior at casual tasks such as question and emotion classification \cite{aldeen2023}. Due to the ambigious and intricate nature of the definition of cultural common sense, human annotation is a solid methodology to evaluate datasets reflectivity of cultural understanding. 

Human annotation can be misleading and unreliable if it is crowd-sourced from non-experts \cite{snow-etal-2008-cheap}. We therefore carefully curate a group of human annotators including domain experts, and provide detailed guidelines when no domain experts are included. The details of annotators and guidelines are given in Appendix \ref{appendix:annotations} and \ref{appendix:annotation_guidelines}.

\subsubsection{LLM-Judge}
In addition to human annotations, we employ three different LLMs as annotators in this study: Llama-3.3-70B-Instruct \cite{llama3-70b-instruct}, Gemma-2-27B-it \cite{gemma2-27b-it}, and GPT-4o. We evaluate them with the same datasets and metrics as those used for human annotators.
We analyze the performance of LLM-Judge for each metric separately and compared with the results of human annotators. This comparison aims to assess the degree to which LLM-Judge could replicate human performance in annotation tasks.

\section{Experiments}

\subsection{Experimental Design}
There are two kinds of experiments in this study. First, we evaluate the quality of benchmark datasets using human annotations. We then repeat the same experiments using generative LLMs instead of human annotators. We compare their performances to understand whether LLM-Judge is competitive to human annotations.

\subsubsection{Human-Centered Experimental Design}
\paragraph{Sampling}

The Central Limit Theorem states that the sampling distribution of the mean will approximate a normal distribution as the sample size increases, regardless of the population's original distribution. The sample size is often context-dependent and depends on the variability within the population. In this study, we sample 100 random instances from each dataset to be annotated. The choice of 100 samples reflects a practical balance between accuracy and computational effort. 

\paragraph{Annotator Selection and Guidelines}

To provide diversity, we assign 31 annotators from different backgrounds. Annotators include undergraduate and graduate students, faculty members, and industry professionals. Each instance is annotated by three annotators and majority voting is applied. Since increasing the annotator count might decrease the agreement monotonically \cite{salminen2021}, we choose to have three annotators. Some annotators are assigned more than one task based on their availability. Depending on the difficulty of the datasets, we assign one week or two weeks to complete a task.

Before starting annotations, all annotators were asked to study a detailed guidelines document, which was written by experts in the language and sociology domain. Annotator guidelines consist of two sections. We first explain datasets in details, and then provide the descriptions of evaluation criterion with sample annotations. The details of the annotators and the guidelines document are given in Appendix \ref{appendix:annotations}.

\paragraph{Inter-annotator Agreement}
Random selection of annotators inherently introduces variability in their interpretations of the assessments of the datasets. Inter-annotator agreement is a crucial metric that quantifies the degree of consensus among multiple annotators. Fleiss's Kappa is an inter-annotator agreement score that measures agreement among multiple annotators.

Fleiss' kappa can produce low values even when there is high observed agreement between raters. This paradox occurs particularly when the observed ratings are skewed towards one or a few categories, and leads to unexpectedly large chance agreement estimates. We therefore use Robust Fleiss' Kappa $K_r$ which provides a more accurate quantification of inter-annotator agreement \cite{falotico2015fleiss}. The details of our approach are given in Appendix \ref{appendix:details_iaa}.

\paragraph{Evaluation Metric}
Majority voting has statistical limitations and lacks accuracy in the multiclass labeling scenario \cite{hernandez2019}. We therefore use the binary labeling where annotators label the datasets using 1 for compliance and 0 for non-compliance to each criteria. The evaluation metric for a criterion is then \emph{Criteria Percentage Accuracy} defined as the total number of positive scores determined by majority voting, divided by the total number of data instances. To satisfy being a high-quality dataset, we set a heuristic threshold of having equal or higher than 90\% of accuracy for all criteria.

\subsubsection{LLM-Judge Experimental Design}
The same strategy presented in the human-centered experimental design (sampling, inter-annotator agreement, and majority voting) is also used in this setup. The only difference is the replacement of human annotators by LLMs.

\paragraph{LLM-Judge Selection}
We employ two open-source LLMs (Llama-3.3-70B-Instruct and Gemma-2-27b-it) and a proprietary LLM (GPT-4o). The reason for choosing the larger models is not to benchmark LLMs against each other but rather to analyze the relationship between LLMs and human annotators. Our aim is to assess in which domains, datasets, and tasks LLMs could potentially replace human annotators or whether it is practical to do so. We use default generation configuration settings for all models. 

\paragraph{LLM-Judge Guidelines}
As human annotators are guided on how to evaluate the dataset quality, we tailor similar guidelines for LLMs, ensuring that they follow the same structured approach as in the human-centered annotations. The evaluation expectations given to human annotators are also shared with the LLMs in the same way \cite{mirzakhmedova2024large}. Annotators are given clear instructions on how to assess the model's performance, and the same structure and prompt.

To have an iterative evaluation process for LLMs, we follow a design where LLMs evaluate each metric separately \cite{bavaresco2024llms}. For each metric, LLMs are prompted individually. For a single dataset, LLM is first asked to evaluate accuracy, followed by other criteria. The model is queried six separate times, once for each metric. This approach allows LLMs to focus on each specific aspect of the data, ensuring that its evaluation of one metric does not influence its judgment of another, and thereby offering a more objective comparison with human annotators. An example of LLM prompt is given in Appendix \ref{appendix:prompt}. We publish the prompts for all datasets in our Github repository. 

\subsection{Experimental Results}

\begin{figure*}[t]
    \centering
    \includegraphics[width=0.8\textwidth]{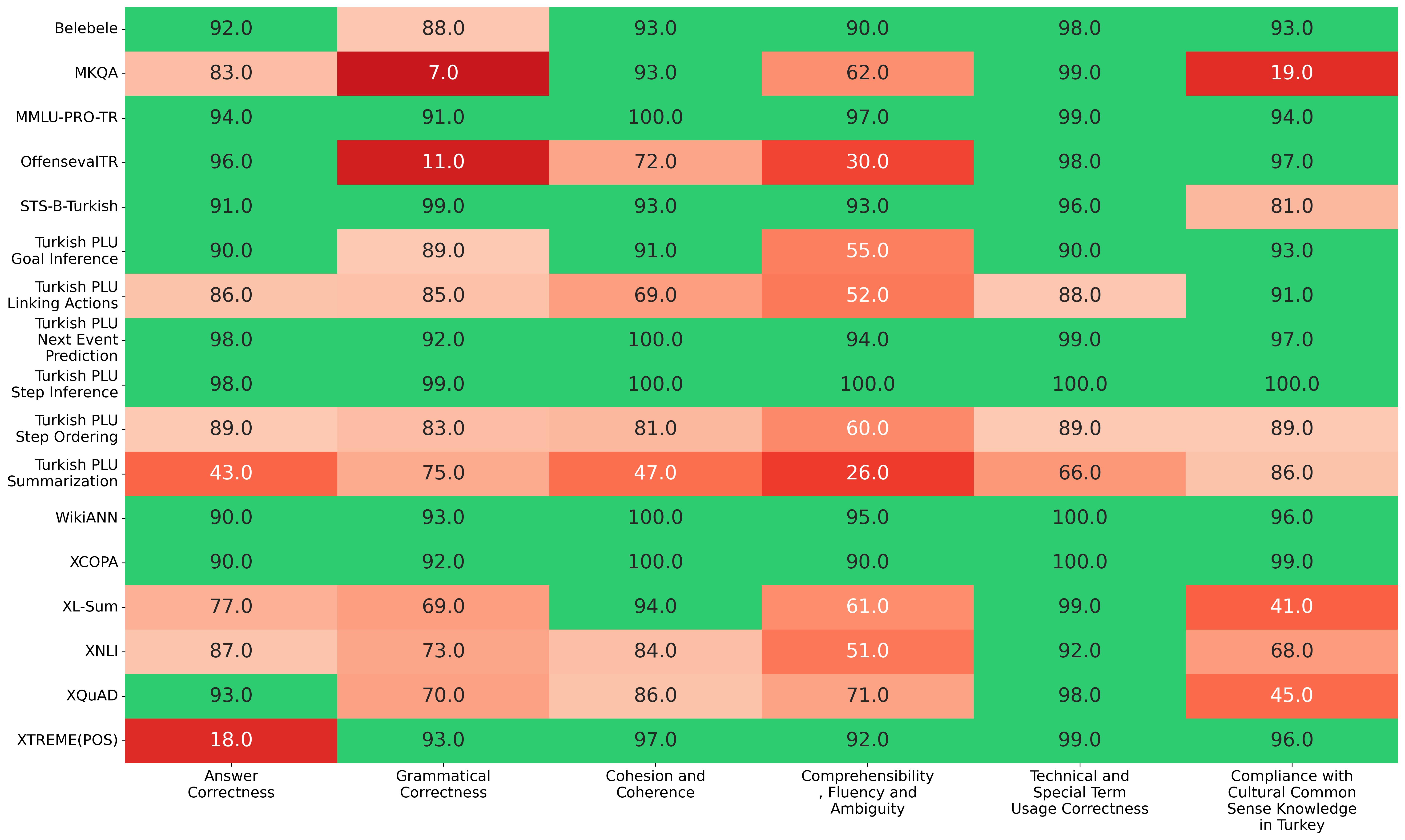}
    \caption{Criteria Percentage Accuracy scores for each dataset (y-axis) across six criteria (x-axis). The cells are colored according to the degree of scores: Positive scores are shades of green, negative ones are shades of red. Scores higher than 90\% are heuristically considered acceptable.}
    \label{fig:results_quality}
\end{figure*}

\subsubsection{Human Annotation Results}

Among 102 evaluations (17 datasets and six criteria), we find that only three of them (WikiANN, XL-Sum, and XNLI) have an agreement score below 0.2. In other terms, 97\% of the experiments have fair or better inter-annotator agreement \cite{landis1977measurement} in this study. The detailed results of inter-annotator agreement are given in Appendix \ref{appendix:iaa}. In Figure \ref{fig:results_quality}, we present the quality evaluation of each dataset examined in this study. 

On the positive side, five of the datasets (MMLU-Pro-Tr, Turkish PLU Next Event Prediction, Turkish PLU Step Inference, WikiANN, and XCOPA) satisfy the criteria of our dataset having higher accuracy than 90\% for all criteria. On the negative side, two of the datasets (Turkish PLU Step Ordering and PLU Summarization) do not satisfy our dataset criteria by having less than 90\% for all criteria. The remaining 10 datasets partially satisfy our dataset criteria. For instance, MKQA and OffensEvalTR have very poor accuracy scores in Grammatical Correctness, and XTREME-POS shows inadequate results in Answer Correctness. Overall, almost 30\% of the benchmark datasets satisfy all criteria, in other terms \emph{70\% of the benchmark datasets fail at our criteria}.

In terms of criteria, only technical and special term usage correctness has more than 90\% in more than 80\% of the datasets examined in this study. That is, \emph{85\% of the criteria are not satisfied by the benchmark datasets}.

\subsubsection{LLM-Judge Results}
We find that 83\% of the experiments have a fair or better annotator agreement when LLM judges are employed. The detailed results of LLM-Judge agreement are given in Appendix \ref{appendix:details_iaa}. 

We notice that LLM judges assign very low scores on the cultural sensitivity of the datasets, while human annotators have relatively higher scores on this criterion. All evaluation scores using only LLM-Judge are provided in Appendix \ref{appendix:results_llmjudge}. Since our ground truth is human annotations, we compare human and LLM-Judge annotations to get any insights on LLM-Judge performance. That is, we analyze whether LLM-Judge can be used as an alternative to human annotations. 

\subsubsection{Human and LLM-Judge Comparison}

\begin{figure*}[t] 
    \centering
    \includegraphics[width=\textwidth]{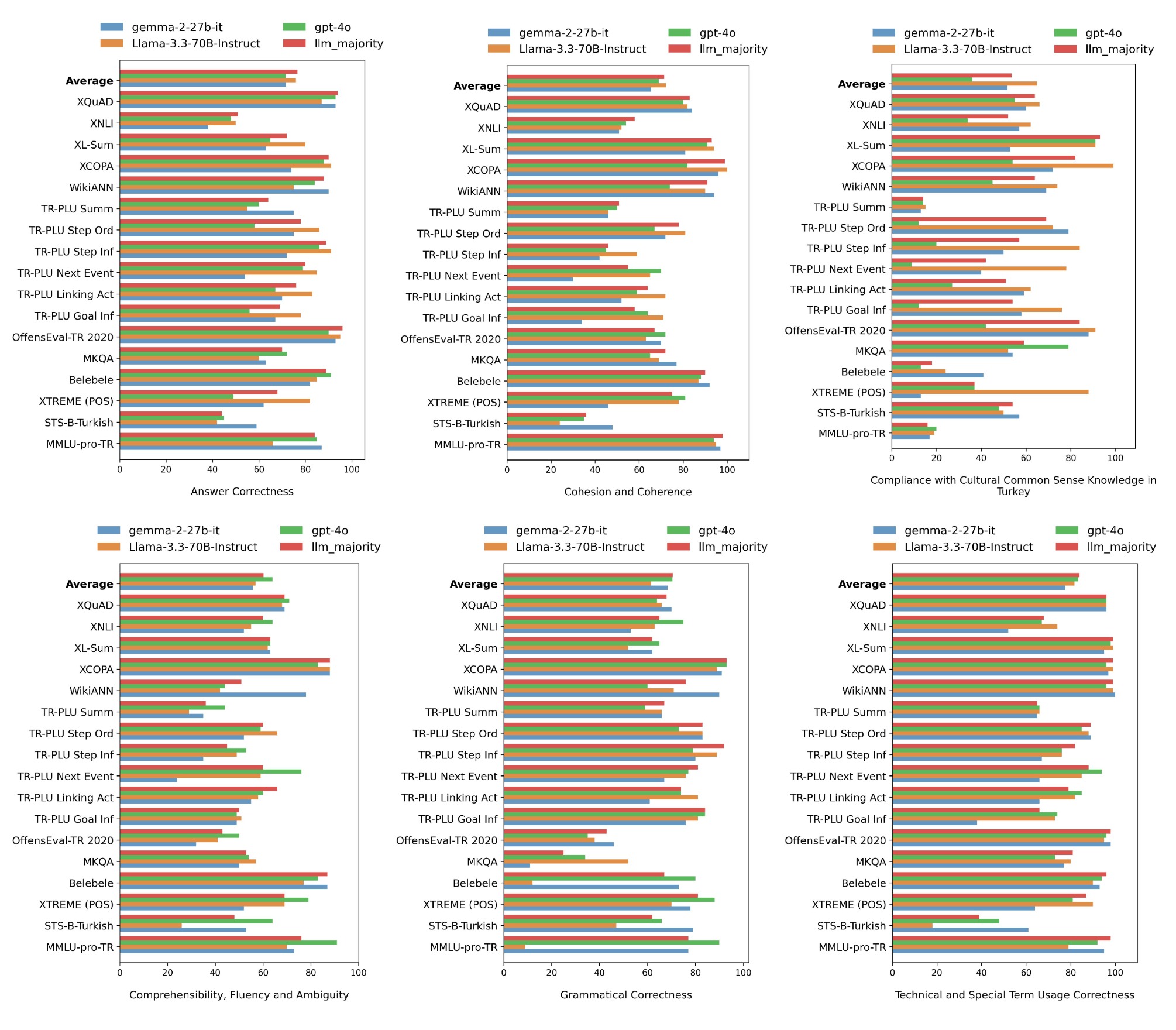} % Figür her iki sütunu kaplasın
    \caption{Comparison between the LLM-Judge majority and human majority labels for each dataset (y-axis) using overlapping ratio (x-axis). The subplots represent an evaluation criterion. The colors represent an LLM model.}
    \label{fig:comparison}
\end{figure*}

\noindent

This section examines how LLMs align with human majority responses. To do so, we calculate \emph{Overlapping Ratio} between LLM-Judge and human annotations to check whether the LLM majority outputs the same answer as the human majority for all datasets. Overlapping ratio is defined as the number of the same annotations/labels provided by LLM-Judge majority and human majority (there are three LLMs and humans in each scenario for labeling each data instance), divided by the total number of data instances annotated. Figure \ref{fig:comparison} shows the results of the overlapping ratio for each dataset. 

\emph{LLM majority have less than 80\% overlapping scores on average with human annotations} for all criteria except technical term usage. Cultural common sense and fluency have the worst overlapping scores among all criteria. This shows that LLM judges are not as good as human annotators, particularly for cultural common sense knowledge and reading fluent and nonambiguous text. 

XCOPA, OffensEval-TR 2020, WikiANN, and XQuAD consistently show high overlapping scores across various criteria. However; STS-B-Turkish, TR-PLU Summ, and XNLI frequently report lower overlapping scores. This shows that LLM judges do not consistently label as humans do in all benchmark datasets. Their labeling capability depends on the characteristics of the dataset.

Llama3.3-70B-Instruct has better overlapping scores than Gemma2-27B and GPT-4o in answer correctness, cohesion and coherence, and cultural common sense knowledge. GPT-4o has better overlapping scores for the remaining. This shows that GPT-4o has a better labeling capability for grammatical and technical tasks, while Llama3.3-70B is good at correctness and cultural knowledge. The discussion of the comparison between LLM-Judge and Human annotations is given in Appendix \ref{appendix:results_comparison}.

\section{Conclusion}
This study evaluated the quality of 17 commonly used Turkish benchmark datasets. Our findings reveal that 70\% of the benchmark datasets fail to meet our criteria, and 85\% of the criteria are not satisfied by these datasets. The successful datasets include MMLU-Pro-Tr, TR-PLU Next Event Prediction, TR-PLU Step Inference, WikiANN, and XCOPA, while the successful criterion is the correctness of technical term usage. These results highlight the need for more rigorous quality control in curating datasets for low-resource languages.

We also considered LLM-Judge annotations as an alternative to human annotations. Our results show that LLM judges are not as effective as human annotators, particularly in understanding cultural common sense knowledge, and interpreting fluent and unambiguous text. In addition, GPT-4o demonstrates stronger labeling capabilities for grammatical and technical tasks, whereas Llama3.3-70B performs better in correctness and cultural knowledge evaluation. In future work, we aim to construct a reliable and high-quality benchmark dataset that addresses the shortcomings identified in this study.

\section{Limitations}
The framework evaluates 17 widely used datasets curated in Turkish language. More datasets, especially those in specialized domains, can be included to reflect more general results. Furthermore, our findings, while significant for Turkish natural language processing, may not be directly transferable to some other low-resource languages. 

The reliance on human annotations introduces potential challenges. Although human evaluators are effective in assessing particular criteria such as cultural common sense, their judgments could be still subjective.

Criteria in the study emphasize linguistic and cultural alignment but may overlook broader notions such as representational biases and regional sensitive topics. The focus of our study on the quality criteria of the data set could also be expanded to consider ethical dimensions.

\section{Ethical Considerations}

Relying on the datasets that fail to meet quality criteria could produce models that are poorly performed for diverse real-world scenarios, particularly in critical domains like healthcare, law, and education. 

Our study highlights the limitations of LLM judges compared to human annotators. There is a risk that future reliance on automated systems for dataset evaluation could compromise the quality of models and systems, particularly for cultural sensitivity or linguistic coherence.

Our experiments on LLM-Judge annotations involve computationally intensive dataset evaluation. There are environmental impacts to consider, given the energy consumption of such processes.

\paragraph{Acknowledgments:} 
We thank to Havelsan for their support at the beginning of this study. We express our sincere gratitude to all annotators who contributed to this study. Special thanks to 
(listed in alphabetical order) 
Abdullah Topraksoy, 
Ahmet Enes Salman, 
Ahmet Kaan Sever, 
Ayşe Aysu Cengiz, 
Başar Yılmaz, 
Çağatay Akpınar, 
Deniz Yılmaz, 
Elif Özge Yılmaz, 
Erdem Orman, 
Esra Darıcı,
Fatih Sinan Esen, 
Görkem Sevinç, 
Güney Kırık, 
İpek Sönmez, 
İsmail Furkan Atasoy
Kaan Engür, 
Kuntay Yılmaz, 
Muhammed İkbal Özbey, 
Mustafa Mert Satılmış, 
Oğuzhan Yusuf Aslanalp, 
Osman Gürlek, 
Saime İpek İşçelebi, 
Sarp Kantar, 
Selçuk Tekgöz, 
Tanay Sütçü, 
Tufan Özkan, 
Yahya Bahadır Karataş, 
Yaren Mercan, 
Yiğit Polat, 
Yusuf Mücahit Çetinkaya, 
Zeynep Berda Akkuş.

\section{Appendix}
\label{sec:appendix}

\subsection{Details of Human Annotations}
\label{appendix:annotations}

The annotators are composed of 31 people from a broad spectrum of backgrounds. The following demographic information is provided by the annotators. The annotators include 24 undergraduate students, two M.Sc.  students, two research assistants, one faculty member, and two industry professionals. The annotators include 24 male and seven female participants. There are 24 participants who are between 20 and 25 years old, and seven participants who are older than 25 years.

\subsection{Details of Annotation Guidelines}
\label{appendix:annotation_guidelines}

The annotator guidelines document aims to guide annotators in their tasks. These guidelines consist of two sections, which are the Common Guideline, and Dataset Specifications. The former is the same for all guidelines. The latter one contains dataset specific information. 

Every annotator is expected to follow the guidelines in order to make the results as much objective and decisive as possible. Depending on the difficulty of the datasets, annotators were assigned one or two weeks to complete the task.

The annotator guidelines document provides detailed explanations with examples of the six criteria outlined in \nameref{sec:criteria}. The document can be found in \href{https://github.com/metunlp/llmevaluation/}{this link}. The document also provides a detailed explanation of the dataset to guide the annotator. This document outlines the column names along with their corresponding definitions and clarifies the specific tasks associated with the dataset. Additionally, it offers an in-depth discussion of the "Answer Correctness" criteria and its relevance within this context

\subsection{Details of Inter-annotator Agreement}
\label{appendix:details_iaa}

Cohen's Kappa measures the agreement between two annotators. Fleiss's Kappa extends Cohen's Kappa to multiple annotators, and Krippendorffs's Alpha additionally handles missing data.

Fleiss' Kappa is given as follows. \\

\[
\kappa = \frac{\bar{P} - \bar{P_e}}{1 - \bar{P_e}}
\]

\[
\bar{P} = \frac{1}{N} \sum_{i=1}^{N} P_i, \quad P_i = \frac{1}{n(n-1)} \sum_{j=1}^{k} n_{ij} (n_{ij} - 1)
\]

\[
\bar{P_e} = \sum_{j=1}^{k} p_j^2, \quad p_j = \frac{\sum_{i=1}^{N} n_{ij}}{Nn}
\]

\text{where:}

\begin{align*}
N & \text{: Total number of samples (100 in our case)} \\
n & \text{: Number of annotators per sample (3 in ours)} \\
k & \text{: Number of labels (0 or 1 in our case)} \\
n_{ij} & \text{: Number of who assign label } j \text{ to sample } i \\
p_j & \text{: Proportion of all assignments to label } j
\end{align*}

\begin{figure*}[ht!]
    \centering
    \includegraphics[width=0.65\textwidth]{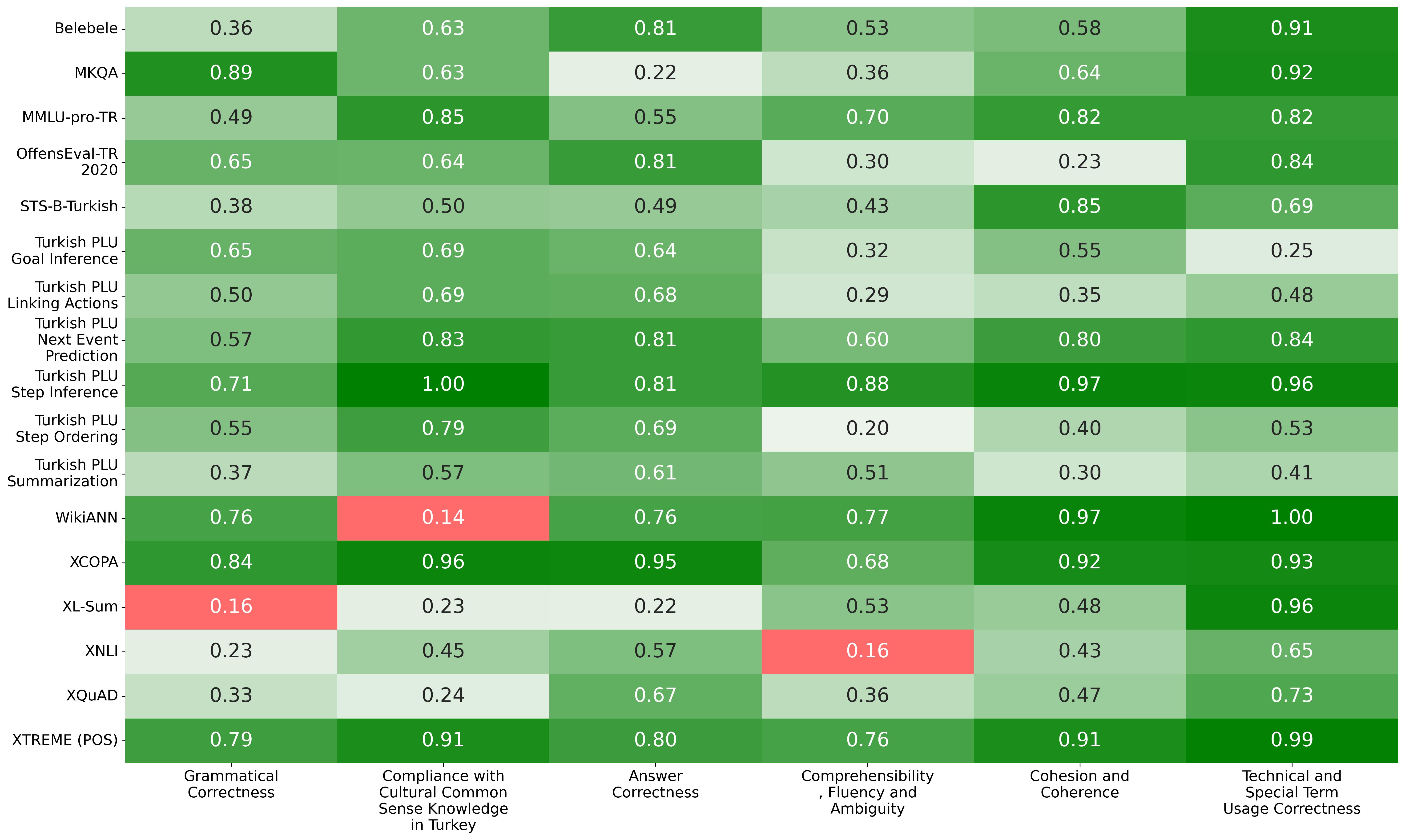}
    \caption{Robust Fleiss' Kappa scores for each dataset annotated by three \textbf{human annotators} across six criteria.}
    \label{fig:results_kappa_human}
\end{figure*}

Although Fleiss' Kappa fits our purpose, there is a drawback to use this metric. It is inconsistent to use Fleiss' Kappa when there is strong agreement among raters. That is, it shows unexpected behavior when there is near-perfect agreement. For example, if annotators vote the same for all entries (perfect agreement), the expected agreement $\bar{P_e}$  would be 1, and the observed agreement $\bar{P}$  would also be 1, which would lead to an undefined value. In a near-perfect agreement situation, $\bar{P_e}$ gets a higher value than $\bar{P}$, and leads to a negative value.

In \cite{falotico2015fleiss}, they proposed a permutation-based method to address this issue. They show that Fleiss’ kappa is inadequate in interpreting high levels of agreement. In addition, they recommend bootstrap techniques for constructing confidence intervals that avoid paradoxes. Their research aligns with our earlier observations.

As we are interested in the agreement of annotators rather than what they voted for here, the proposed solution involves generating permutations of category frequencies for each row of the data table, substituting the original vectors with these permutations, and recalculating Fleiss' kappa. By repeating this process C times and summarizing the resulting kappa values using a robust statistic like the median, the authors derive a new measure, Robust Fleiss' Kappa $K_r$ which provides a more accurate quantification of inter-annotator agreement. 

In our experiments, we set C, the number of permutations, to 100. For each permutation, we calculated the Fleiss' Kappa based on the permuted score combinations and then averaged these values, following the method outlined in the paper.

To compute the confidence intervals, we again used the methods explained in the paper. We generate bootstrap samples from the original voting matrix by randomly sampling rows with replacements. For each bootstrap sample, we calculate the Robust Fleiss' Kappa. This process is repeated B times (with B=1000 in our experiments), resulting in B values of Robust Fleiss' Kappa. Using a confidence level of $1 - \alpha$ = 0.95 (95$ \% $), we determine the bounds of the confidence interval based on these B values. This implies that there is a 95$ \%$ likelihood that the true inter-annotator agreement value lies within the confidence intervals reported in the tables. 

To ensure the success of Robust Fleiss' Kappa in our research, we aim for an agreement score higher than 0.2, which is considered a fair level of agreement, as shown in Table \ref{tab:fleiss_kappa}.

\begin{table}[h]
    \centering
    \small
    \begin{tabular}{|c|l|}
        \hline
        \textbf{Fleiss' Kappa} & \textbf{Interpretation} \\ \hline
        $\kappa < 0$ & Poor agreement \\ \hline
        $0.00 - 0.20$ & Slight agreement \\ \hline
        $0.21 - 0.40$ & Fair agreement \\ \hline
        $0.41 - 0.60$ & Moderate agreement \\ \hline
        $0.61 - 0.80$ & Substantial agreement \\ \hline
        $0.81 - 1.00$ & Almost perfect agreement \\ \hline
    \end{tabular}
    \caption{Interpretation of Fleiss' Kappa Values according to \citet{landis1977measurement}.}
    \label{tab:fleiss_kappa}
\end{table}

\subsection{Sample LLM-Judge Prompt}
\label{appendix:prompt}

A sample prompt for the MMLUPro-TR dataset with the accuracy metric is given as follows (English translations are given in parantheses).

\textbf{Veri kümesi sütunları} \textit{(Dataset Columns)}:

\textbf{question\_id}: Soruya özel numara \textit{(Unique identifier for each question)}.

\textbf{question}: Soru metni \textit{(Text of the question)}.

\textbf{options}: On adet cevap şıkkı \textit{(Ten answer choices)}.

\textbf{answer}: Doğru cevabın İngilizce alfabede karşılık geldiği harf \textit{(The letter corresponding to the correct answer in the English alphabet)}. (örn: 1. şık → A, 4. şık → D) \textit{(e.g., choice 1 → A, choice 4 → D)}.

\textbf{answer\_index}: Doğru cevabın listedeki indisi \textit{(The index of the correct answer in the list)}. (indisler 0’dan başlıyor; \textit{indices start from 0)}.

\textbf{category}: Sorunun gerektirdiği bilginin alanı \textit{(The field of knowledge required by the question)}.

\textbf{src}: Kaynak \textit{(Source)}.

\textbf{Değerlendirme sütunları} \textit{(Evaluation Columns)}:

\textbf{Doğruluk (Accuracy)}: Aşağıdaki iki soruya da cevabınız “evet” ise kutuya \texttt{1} yazın, birine bile cevabınız hayır ise \texttt{0} yazın.\\
\textit{(If your answer to both of the following questions is “yes,” write \texttt{1} in the box. If the answer is "no" for one question,” write \texttt{0})}.

\begin{itemize}
    \item \textbf{a.} Doğru cevap şıklarda var mı? \textit{(Is the correct answer among the options?)}
    \item \textbf{b.} Soru için verilen cevap şıkkı doğru şık mı? \textit{(Is the selected option the correct answer for the question?)}
\end{itemize}

\subsection{Detailed Results of Inter-Annotator Agreement}
\label{appendix:iaa}

In Figure \ref{fig:results_kappa_human}, we present Robust Fleiss' Kappa Scores among three human annotators who labeled each dataset based on six criteria. The y-axis represents different datasets, and the x-axis represents our six criteria. The $cell(i,j)$ represents the Fleiss' Kappa score of $dataset_j$ for $criteria_i$. Since a score of 0.2 or higher is considered fair agreement, we accept this as a sufficient threshold in our study. All green values in the table has thereby scores above 0.2. For the WikiANN, XL-Sum, and XNLI datasets; there is a single criterion where the agreement score falls below 0.2. 

\begin{figure*}[ht!]
    \centering
    \includegraphics[trim={3.5cm 0cm 14cm 0},clip,width=0.65\textwidth]{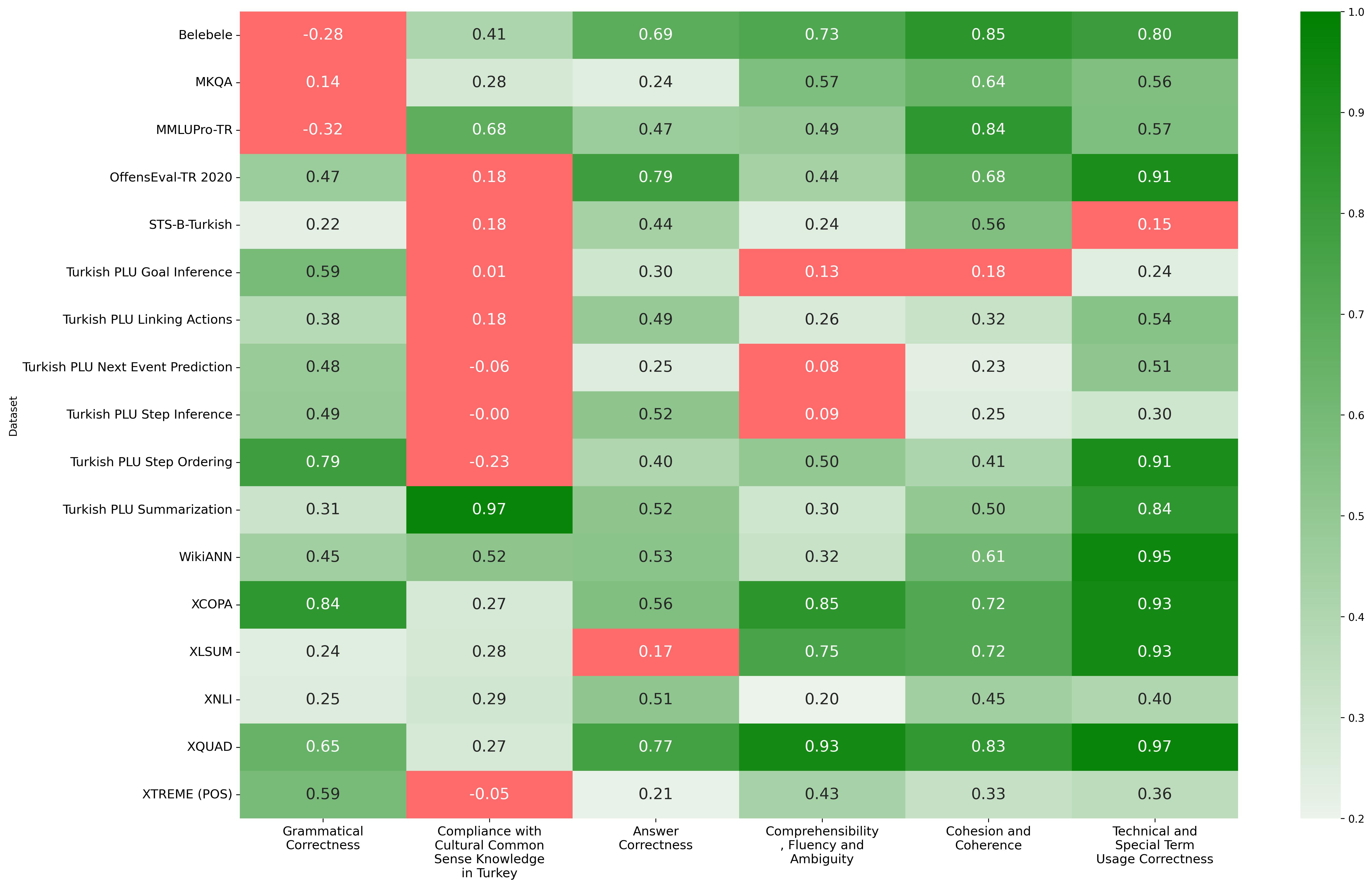}
    \caption{Robust Fleiss' Kappa scores for each dataset annotated by three \textbf{LLM-Judge annotators} across six criteria.}
    \label{fig:results_kappa_llm}
\end{figure*}

In Figure \ref{fig:results_kappa_llm}, we present Robust Fleiss' Kappa Scores among three LLM-Judge annotators. The number of agreement scores are mostly below 0.2 in cultural common sense knowledge. The inter-annotator agreement between LLM-Judge models is worse than the one between human annotators, since 17 comparisons have below 0.2 score in LLM-Judge while this number is only three comparisons in human annotations. In other terms, 17 out of 102 experiments (17\%) have poor agreement.

\subsection{Detailed Results of LLM-Judge}
\label{appendix:results_llmjudge}

\begin{figure*}[h]
    \centering
    \includegraphics[width=0.7\textwidth]{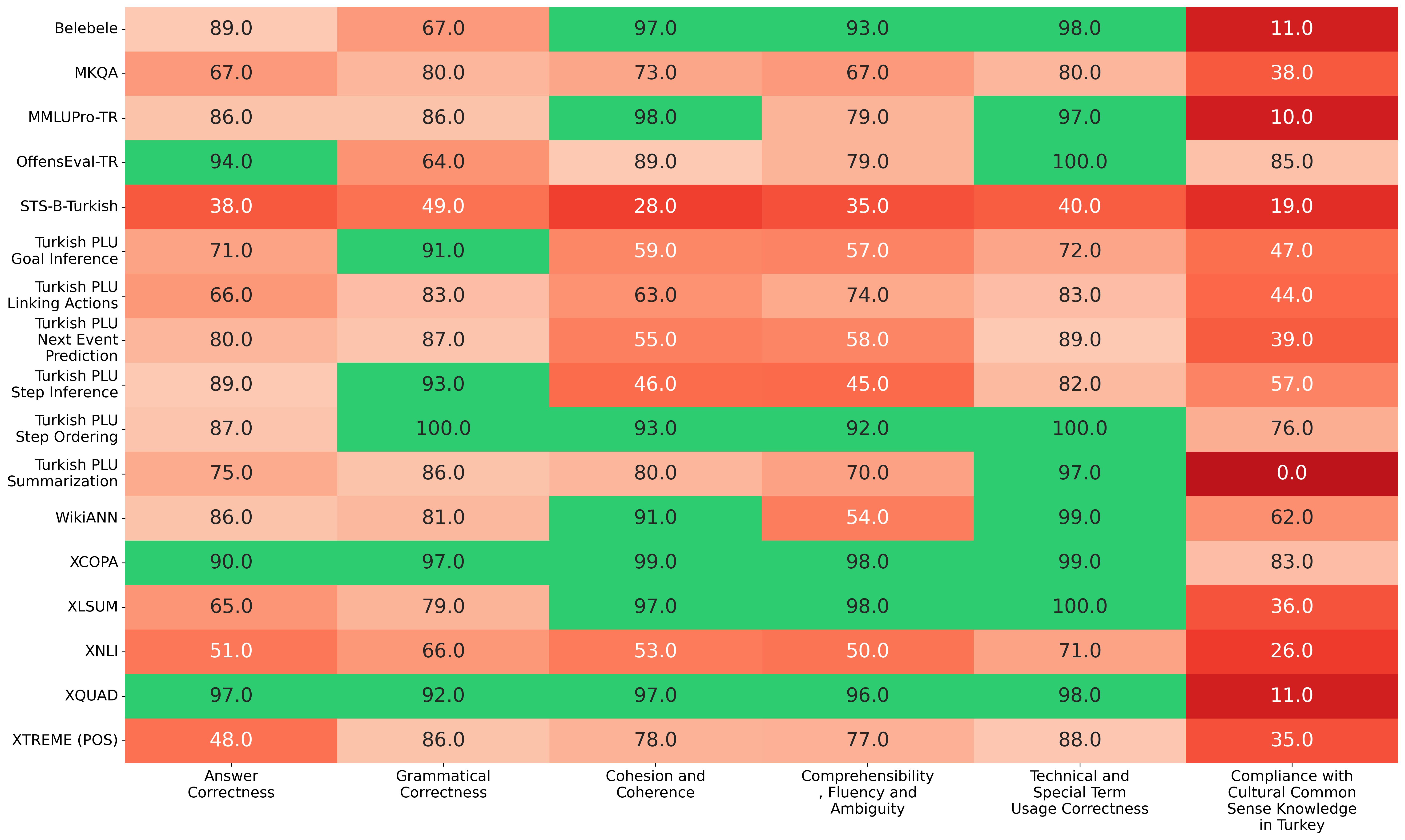}
    \caption{Criteria Percentage Accuracy scores (Majority Voting) for each dataset annotated by three LLMs Across six criteria.}
    \label{fig:results_llm}
\end{figure*}

In Figure \ref{fig:results_llm}, we present LLM-Judge evaluation results. The results show that LLMs perform consistently well in the datasets such as XQUAD, Belebele, and Turkish PLU Step Ordering, especially in the metrics such as Accuracy and Grammar Correctness. For instance, in the XQUAD dataset, high scores were achieved in answer accuracy (97\%), grammar correctness (92\%), and technical term usage (98\%). This suggests that LLMs are aligned with evaluators and handle technical aspects of language well. Similar consistency is seen in grammar and technical term usage in the Belebele and Turkish PLU Step Ordering datasets.

However, the result also reveals inconsistencies in datasets such as STS-B Turkish and XTREME (POS), particularly in the metrics including fluency, contextual understanding, and cultural knowledge. In the STS-B Turkish dataset, low scores in answer accuracy (38\%) and contextual alignment (28\%) suggest that the model struggles with these tasks. In XTREME (POS), although grammar accuracy is high (86\%), performance drops in more challenging metrics like cultural alignment and fluency (35\%).

Overall, the results indicate that LLMs perform well in technical accuracy and grammar-focused metrics but show inconsistencies in tasks requiring natural language flow, contextual understanding, and cultural awareness. This can suggest that while models excel in certain tasks, they still have room for improvement in more complex and context-driven tasks.

\subsection{Detailed Results of Comparison between LLM and Human Labels}
\label{appendix:results_comparison}

\textbf{LLM Majority} typically demonstrates high accuracy, particularly on datasets like OffensEval-TR 2020 (96\%) and XQuAD (94\%). However, its performance drops on certain datasets such as STS-B-Turkish (44\%) and XNLI (51\%). The model’s consistency varies depending on the dataset; for example, it shows strong agreement on XCOPA (99\%) and Belebele (90\%), but weak consistency on STS-B-Turkish (36\%) and TR-PLU Summ (51\%). In terms of cultural sensitivity, the model excels on XL-Sum (93\%) and OffensEval-TR 2020 (84\%), but falls short on TR-PLU Summ (14\%) and MMLU-pro-TR (16\%). For metrics like comprehensibility, fluency, and ambiguity, the model performs well on datasets like XCOPA (88\%) and Belebele (87\%), but faces challenges on TR-PLU Summ (36\%) and TR-PLU Step Inf (45\%). Grammatical accuracy is strong on XCOPA (93\%) and TR-PLU Step Inf (92\%), but problematic on MKQA (25\%) and OffensEval-TR 2020 (43\%). Technical terminology is well-handled on WikiANN (99\%) and XCOPA (99\%), but more challenging on STS-B-Turkish (39\%) and TR-PLU Summ (65\%).

\textbf{GPT-4o} performs exceptionally on datasets such as MMLU-pro-TR (85\%) and XQuAD (93\%). However, its performance lags on datasets like TR-PLU Step Ord (58\%) and XNLI (48\%). The model's consistency is solid on datasets like XCOPA (82\%) and XL-Sum (91\%), but weak on STS-B-Turkish (35\%) and TR-PLU Step Inf (45\%). In terms of cultural sensitivity, it excels on XL-Sum (91\%) and MKQA (79\%), but underperforms on TR-PLU Next Event (9\%) and TR-PLU Goal Inf (12\%). For comprehensibility and fluency, the model shows strong performance on MMLUPro-TR (91\%) and XCOPA (83\%), but experiences ambiguity on TR-PLU Summ (44\%) and TR-PLU Step Inf (53\%). Grammatical accuracy is high on XCOPA (93\%) and MMLU-pro-TR (90\%), but weak on MKQA (34\%) and OffensEval-TR 2020 (35\%). Technical terminology is well-managed on WikiANN (96\%) and XCOPA (96\%), but lacks precision on TR-PLU Goal Inf (74\%) and TR-PLU Step Inf (76\%).

\textbf{Llama} achieves its best results on OffensEval-TR 2020 (95\%) and TR-PLU Step Inf (91\%), but performs poorly on datasets like STS-B-Turkish (42\%) and XNLI (50\%). Its consistency is strong on XCOPA (99\%) and XL-Sum (94\%), but inconsistent on STS-B-Turkish (24\%) and TR-PLU Summ (46\%). In terms of cultural sensitivity, it performs well on XCOPA (99\%) and OffensEval-TR 2020 (91\%), but struggles with TR-PLU Summ (15\%) and MMLU-pro-TR (19\%). For comprehensibility and fluency, it excels on XCOPA (88\%) and MMLUPro-TR (70\%), but faces ambiguity on TR-PLU Summ (29\%) and TR-PLU Next Event (59\%). Grammatical accuracy is high on XCOPA (89\%) and TR-PLU Step Inf (89\%), but poor on MMLU-pro-TR (9\%) and Belebele (12\%). Technical terminology is well-handled on WikiANN (99\%) and XCOPA (99\%), but problematic on STS-B-Turkish (18\%) and TR-PLU Goal Inf (73\%).

\textbf{Gemma} performs well on datasets like XCOPA (96\%) and TR-PLU Step Ord (79\%), but struggles on TR-PLU Next Event (24\%) and TR-PLU Step Inf (35\%). Its consistency is strong on XCOPA (96\%) and Belebele (87\%), but weak on TR-PLU Next Event (24\%) and TR-PLU Step Inf (35\%). In terms of cultural sensitivity, it excels on OffensEval-TR 2020 (88\%) and TR-PLU Step Ord (79\%), but underperforms on TR-PLU Summ (13\%) and XTREME (POS) (13\%). For comprehensibility and fluency, it performs well on XCOPA (88\%) and Belebele (87\%), but shows ambiguity on TR-PLU Next Event (24\%) and TR-PLU Step Inf (35\%). Grammatical accuracy is excellent on WikiANN (99\%) and XCOPA (91\%), but low on MKQA (11\%) and XTREME (POS) (13\%). Technical terminology is handled well on WikiANN (99\%) and XCOPA (97\%), but lacks precision on TR-PLU Goal Inf (38\%) and TR-PLU Next Event (66\%).

\bibliography{anthology,custom}
\bibliographystyle{acl_natbib}

\end{document}